\documentclass[sigconf]{acmart}

\usepackage{graphicx}
\usepackage{multirow}
\usepackage{algorithm}
\usepackage{algcompatible}
\usepackage{subcaption}

\AtBeginDocument{%
  \providecommand\BibTeX{{%
    \normalfont B\kern-0.5em{\scshape i\kern-0.25em b}\kern-0.8em\TeX}}}

\copyrightyear{2020}
\acmYear{2020}
\setcopyright{acmcopyright}\acmConference[MM '20]{Proceedings of the 28th ACM International Conference on Multimedia}{October 12--16, 2020}{Seattle, WA, USA}
\acmBooktitle{Proceedings of the 28th ACM International Conference on Multimedia (MM '20), October 12--16, 2020, Seattle, WA, USA}
\acmPrice{15.00}
\acmDOI{10.1145/3394171.3413952}
\acmISBN{978-1-4503-7988-5/20/10}

\begin{document}
\fancyhead{}

\title{Webly Supervised Image Classification with Metadata: Automatic Noisy Label Correction via Visual-Semantic Graph}
    \author{Jingkang Yang}
    \authornote{Equal Contribution. Work done during an internship at SenseTime EIG Research.}
    \affiliation{\institution{SenseTime Research}
    \institution{Dept. of Electrical and Computer Engineering, Rice University}}
    \email{yangjingkang@sensetime.com}
    
    \author{Weirong Chen}
    \authornotemark[1]
    \affiliation{\institution{SenseTime Research}
    \institution{Dept. of Comp. Sci. \& Eng., The Chinese University of Hong Kong}}
    \email{chenweirong@sensetime.com}
    
    \author{Litong Feng}
    \affiliation{\institution{SenseTime Research}}
    \email{fenglitong@sensetime.com}
    
    \author{Xiaopeng Yan}
    \affiliation{\institution{SenseTime Research}}
    \email{yanxiaopeng@sensetime.com}
    
    \author{Huabin Zheng}
    \affiliation{\institution{SenseTime Research}}
    \email{zhenghuabin@sensetime.com}
    
    \author{Wayne Zhang}
    \affiliation{\institution{SenseTime Research}
    \institution{Qing Yuan Research Institute, Shanghai Jiao Tong University}}
    \email{wayne.zhang@sensetime.com}

\renewcommand{\shortauthors}{J. Yang and W. Chen, et al.}

\begin{abstract}
    Webly supervised learning becomes attractive recently for its efficiency in data expansion without expensive human labeling. However, adopting search queries or hashtags as web labels of images for training brings massive noise that degrades the performance of DNNs.
    Especially, due to the semantic confusion of query words, the images retrieved by one query may contain tremendous images belonging to other concepts.
    For example, searching `tiger cat' on Flickr will return a dominating number of tiger images rather than the cat images.
    These realistic noisy samples usually have clear visual semantic clusters in the visual space that mislead DNNs from learning accurate semantic labels.
    To correct real-world noisy labels, expensive human annotations seem indispensable.
    Fortunately, we find that metadata can provide extra knowledge to discover clean web labels in a labor-free fashion, making it feasible to automatically provide correct semantic guidance among the massive label-noisy web data. In this paper, we propose an automatic label corrector \emph{VSGraph-LC} based on the visual-semantic graph. \emph{VSGraph-LC} starts from anchor selection referring to the semantic similarity between metadata and correct label concepts, and then propagates correct labels from anchors on a visual graph using graph neural network (GNN).
    Experiments on realistic webly supervised learning datasets Webvision-1000 and NUS-81-Web show the effectiveness and robustness of \emph{VSGraph-LC}. Moreover, \emph{VSGraph-LC} reveals its advantage on the open-set validation set.
\end{abstract}




\maketitle

\section{Introduction}
\label{sec:introduction}
Deep convolutional neural networks (CNNs) are successful by virtue of large-scale datasets with human annotation~\cite{lecun2015deep}. However, human annotation is extremely time-consuming and expensive, which impedes the further expansion of those big datasets~\cite{waldrop2019news}. 
To overcome this limitation, researchers use web crawlers to collect billions of images and annotate them directly using text queries or hashtags~\cite{mahajan2018instagram,krasin2017openimages}. 
However, due to the ambiguity or polysemy of the query fed into the search engine, label noise is subsequently introduced.
Therefore, webly supervised learning, aiming at using huge scalable web-crawled data directly for networks training by suppressing label noise, has attracted great attention recently~\cite{algan2019image}.

Early exploration in this direction relies on human-verified clean subsets. Representative methods using clean subsets include MentorNet~\cite{jiang2017mentornet} and CleanNet~\cite{lee2018cleannet}.
However, with the trend of rapid growth in the size of webly datasets, building clean subsets with manual verification becomes more infeasible, especially when the number of categories exceeds ten thousands~\cite{wu2019tencent}.
Therefore, recent works prefer models without clean subset dependencies, making webly supervised learning fully automatic. 
To this end, some works strengthen networks' endurability against label noise using moving average of model predictions~\cite{tanaka2018joint}, 
loss function modification~\cite{manwani2013noise,ghosh2017robust} or
regularization~\cite{zhang2017mixup,miyato2018virtual} .
Co-teaching uses two different networks to mutually detect label noise, doubly ensuring the model's denoising ability~\cite{han2018co}. 
Other works identify noisy samples based on some hypotheses including data density~\cite{guo2018curriculumnet,han2019deep} or model confidence~\cite{yang2020webly}.

Although the aforementioned methods effectively enhance the model against label noise, especially for outliers, suppressing \textit{semantic label noise} is critical but untouched. To clarify, semantic label noise exists due to query's polysemy or insufficient semantic resolution. Usually, semantic label noise would be severe in some category, which is composed of a large number of samples reflecting another  semantic concept.
As webly datasets come from real-world, those off-target semantics are usually out-of-distribution~(i.e., deviate from all semantic labels or out of interests in the test sets).

\begin{figure}[!t]
	\label{fig:intro}
	\centering
	\begin{subfigure}{.46\textwidth}
		\centering
		\includegraphics[width=\linewidth]{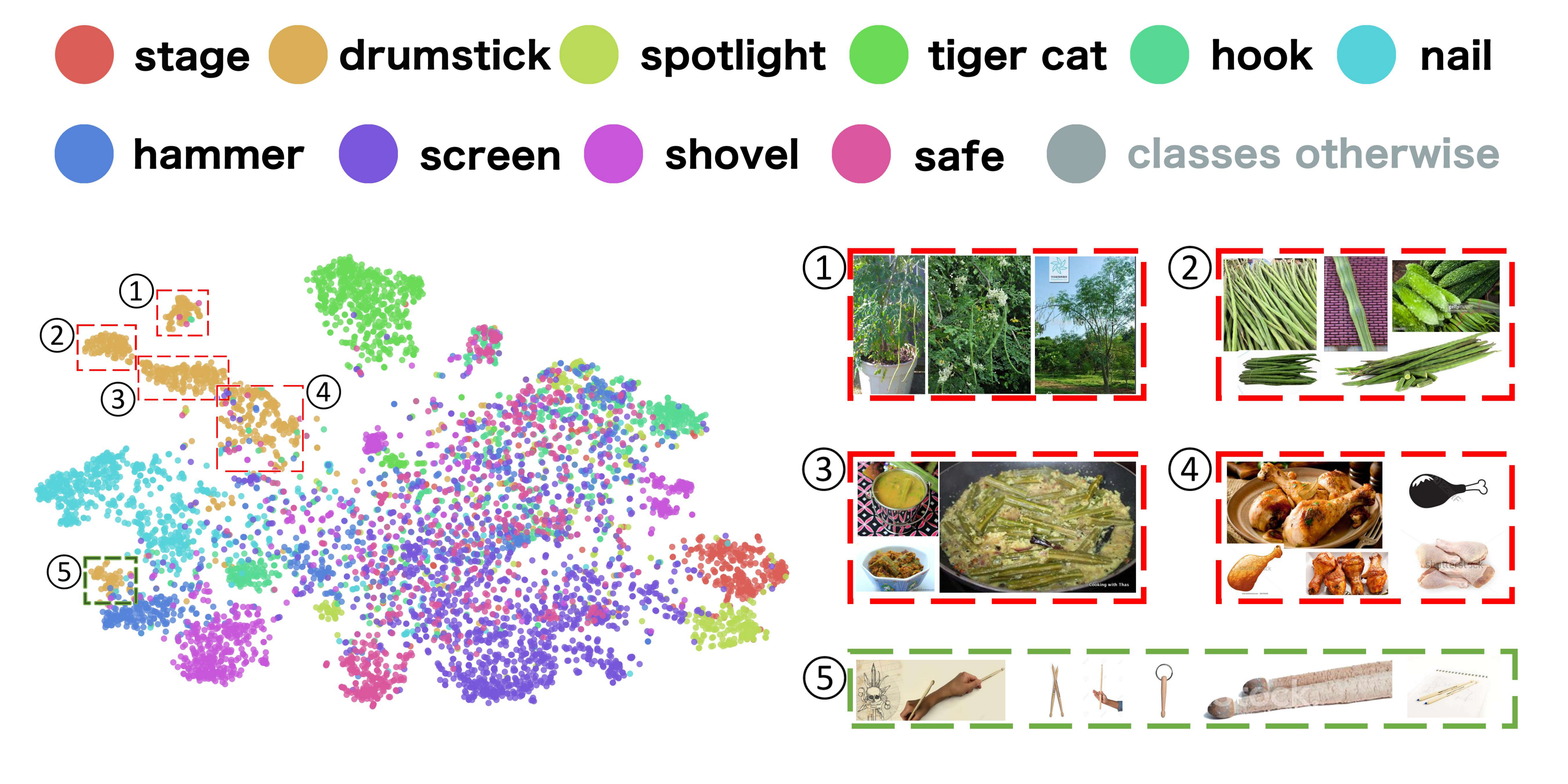}  
		\caption{Colors reflect web labels. Web label 'Drumstick' shows representative images corresponding to 5 regions of interest. Only Region-5 corresponds to the correct concept}
		\Description{Web labels on t-SNE map}
		\label{fig:intro-1}
	\end{subfigure}
	
	\begin{subfigure}{.23\textwidth}
		\centering
		\includegraphics[width=\linewidth]{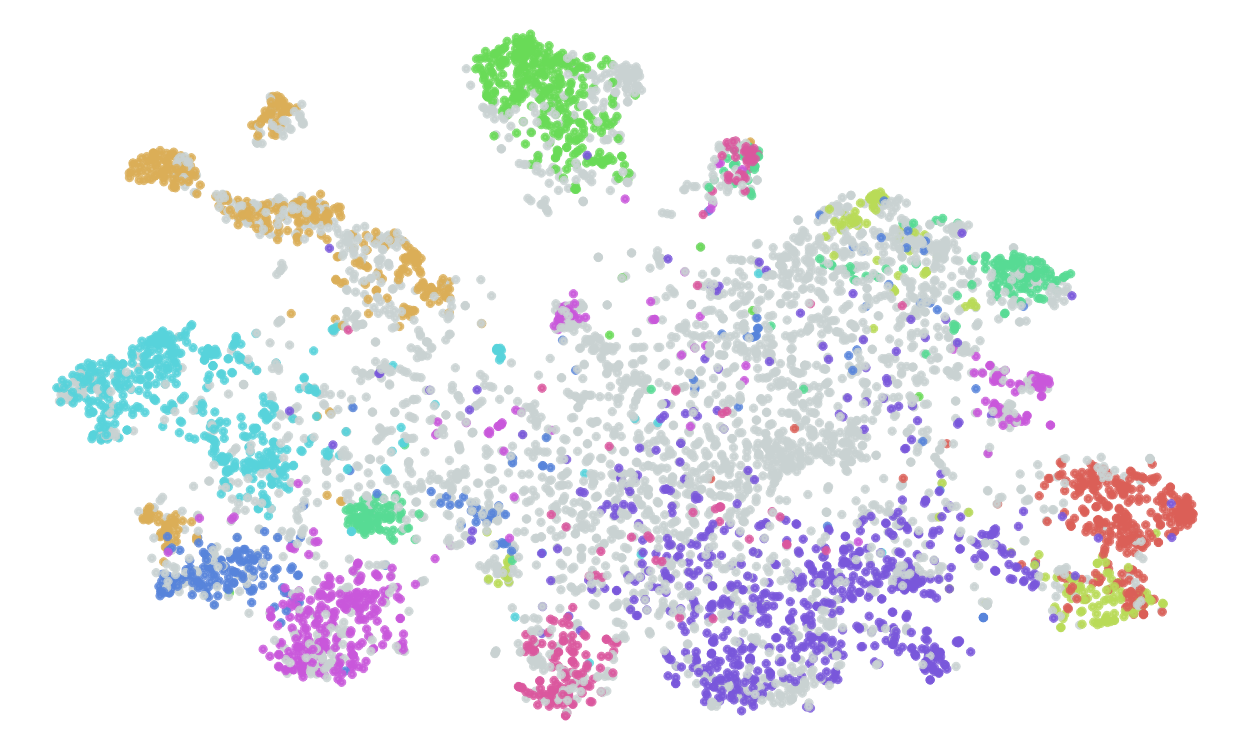}  
		\caption{Prediction by Co-teaching}
		\Description{Prediction by Co-teaching}
		\label{fig:intro-coteaching}
	\end{subfigure}
	\begin{subfigure}{.23\textwidth}
		\centering
		\includegraphics[width=\linewidth]{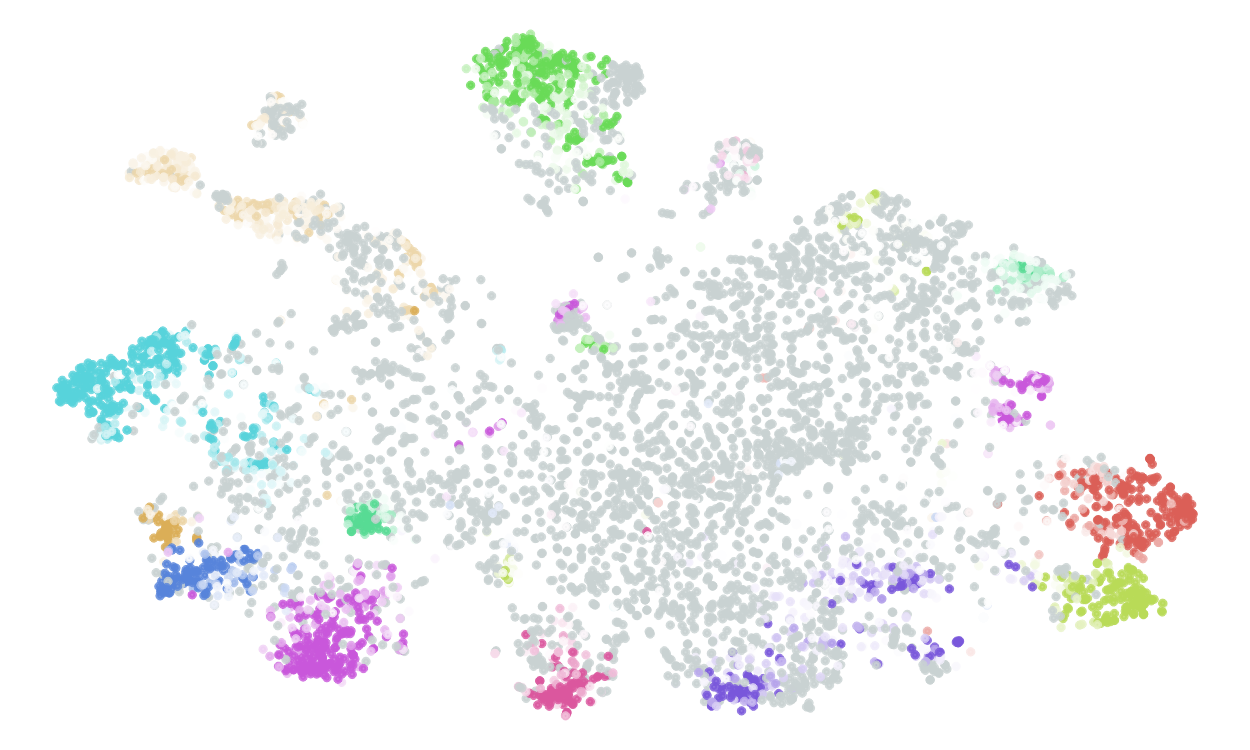}  
		\caption{Prediction by VSGraph-LC}
		\Description{Prediction by VSGraph-LC}
		\label{fig:intro-ours}
	\end{subfigure}
	
	\caption{T-SNE visualization~\cite{maaten2008visualizing} of WebVision-pretrained ResNet50~\cite{he2016deep} features on 10 selected categories. Three observations are highlighted: (1)~CNN models that trained from WebVision can distinguish different semantics within a category, even when semantics mismatch category definition. (2)~Severe semantic label noise is a real-world problem, as majority images of class `drumsticks' deviate the true concept of percussion mallets. (3)~Co-teaching fails to correct the majority semantic label noise, but our VSGraph-LC is able to. Node brightness represents prediction confidence}
	\Description{T-SNE visualization of WebVision-pretrained ResNet50 features on 10 selected categories for web labels and model predictions}
	\label{fig:intro_tsne}
\end{figure}

For example, Figure~\ref{fig:intro-1} shows that although web label `drumsticks' has its correct semantic concept of `percussion mallets' according to the test set, the majority of training samples actually belong to concepts of `drumstick trees/vegetable' and `chicken legs' due to polysemy. These out-of-distribution noisy samples clearly cluster themselves in the visual feature space. 
As a result, density assumption popularly adopted by previous methods~\cite{guo2018curriculumnet,han2019deep} will regard `chicken legs' and `drumstick trees/vegetable' samples falsely as clean data. Figure~\ref{fig:intro-coteaching} further shows the ineffectiveness of the representative self-training method Co-teaching, where most samples belong to off-target semantics are still predicted as positive.

\begin{figure}
	\centering
	\includegraphics[width=\linewidth]{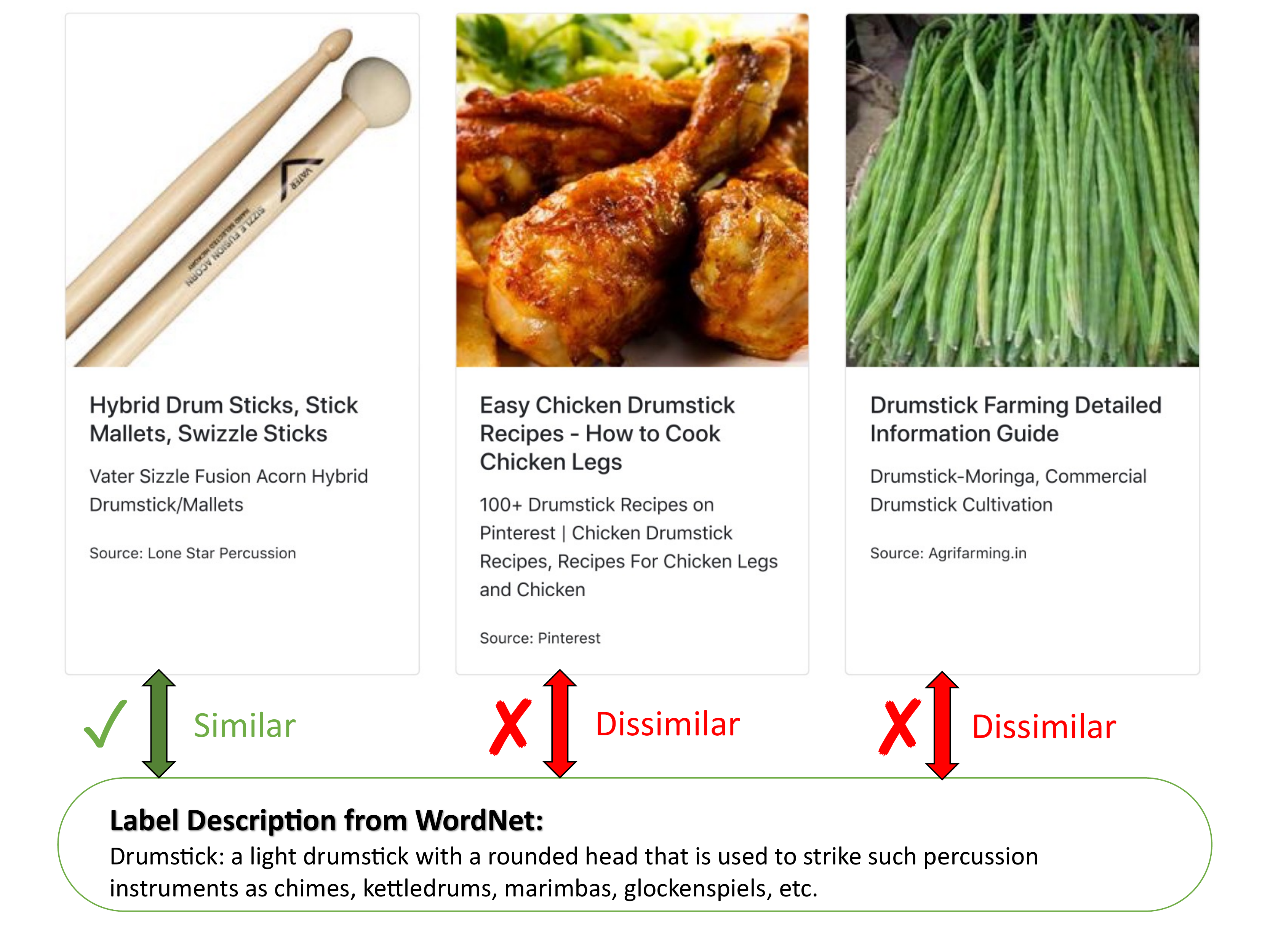}
	\caption{Exemplar metadata associated with web images labeled as `Drumstick'. 
	By comparing metadata with label description, images with \textit{semantic label noise} will be detected}
	\Description{Exemplar metadata associated to web images with label `Drumstick'}
	\label{fig:text}
\end{figure}

As webly dataset is crawled from the Internet, text metadata associated with web images 
has great potential to provide valuable information, which, however, has been ignored for a long time. In this paper, we aim to take advantage of extra knowledge provided by metadata to suppress label noise, especially semantic label noise.
Figure~\ref{fig:text} shows that information in metadata and label description can be used to detect semantic label noise. By using off-the-shelf natural language processing (NLP) models~\cite{yang2019xlnet}, we convert the comparison between label description and metadata from human cognition level to text feature space, which ensures a fully automated process to precisely locate samples with correct semantic concepts, without the need for expensive manual annotations.
	
Hence, we are motivated to build an automatic pipeline for webly supervised learning with metadata. Specifically, we propose a label corrector named VSGraph-LC, which first selects anchor samples for each category through matching label description from WordNet~\cite{miller1998wordnet} and metadata of every crawled image using a powerful NLP model XLNet~\cite{yang2019xlnet}. To help those semantically correct anchors propagate their web labels towards more samples, we leverage a graph neural network (GNN)~\cite{kipf2016gcn} training on $k$-NN visual feature graph of the entire training set.
The corrected labels substitute the former noisy web labels 
for finetuning our final model.

In summary, our contributions are mainly three-fold:
\begin{itemize}
    \item We explore two understudied but important factors under webly supervised learning setting: semantic label noise and text metadata.
    \item A human-labor-free label correcting framework that fully exploits the merits of GNN and CNN is proposed as VSGraph-LC, ignited by anchors automatically selected by metadata.
    \item The proposed framework is shown effective on NUS-81-Web and WebVision datasets, and reaches the state-of-the-art result on WebVision-1000. VSGraph-LC produces more appealing results if the test set contains out-of-distribution images\footnote{Known as open-set recognition task, introduced in~\cite{kuznetsova2018open}}.
\end{itemize}

\section{Related Work}
\label{sec:related}
\subsection{Webly Supervised Learning}
\label{sec:related-webly}
A formal definition of webly supervised learning is in Section~\ref{S:problem}.
Based on the dependency of clean subsets, webly supervised learning methods can be divided into two categories.
Methods that use clean subsets to guide the detection of label noise include MentorNet, which learns a dynamic curriculum as a sample-weighting scheme from a human-labeled subset~\cite{jiang2017mentornet}. CleanNet transfers knowledge learned from human-verified samples from a fraction of categories to the entire dataset for sample reweighting~\cite{lee2018cleannet}. In~\cite{xiao2015learning}, a probabilistic graphical model is learned from a human-verified subset to depict relationships between images, class labels, and label noise. However, with the trend that webly datasets are exceeding billions of training data with more than ten thousand categories~\cite{wu2019tencent}, the exponential increase of human annotations seems infeasible.

Therefore, solving webly supervised image classification without any human-verified labels attracts more attention recently. CurriculumNet trains an image classifier with a curriculum arranged in an unsupervised manner~\cite{guo2018curriculumnet}. By assuming that samples from high-density regions in visual feature space have more correct labels, a three-stage training strategy is designed to feed model from clean to noisy samples. 
A similar assumption is adopted in~\cite{han2019deep}, which selects prototypes in high-density regions. All samples get their labels corrected based on the similarity between samples and prototypes. 
Co-teaching trains two networks simultaneously, letting one be trained on possible clean samples selected by the other. Such cross-update can reduce the self-accumulative error from single model and therefore enhances robustness.
However, according to our inspection in Section~\ref{sec:introduction} and Figure~\ref{fig:intro_tsne}, previous annotation-free methods are vulnerable to massive semantic label noise. 
\cite{shah2019inferring} creatively leverages text from a strong pretrained phrase generator to suppress label confusion, but requires a complicated two-stream design with large network architecture.
Thus, we are motivated to propose a fully automatic pipeline to solve the semantic label noise problem, aided by metadata crawled with web images.

\subsection{Metadata and Concept Learning}
\label{sec:related-text}
Online platforms, including search engines and social media, can not only provide abundant web images, but also meaningful metadata. Various computer vision tasks utilize the potential value of metadata, including powerful visual-language models pretraining~\cite{lu2019vilbert, su2019vlbert}, image retrieval~\cite{espinoza2013earth,li2015weakly}, and visual question answering~\cite{krishna2017visualgenome,sharma2018conceptual}. 
However, for the problem of webly supervised learning, metadata is unfortunately neglected.
    
Recent work has also proposed several visual concept discovery approaches based on metadata.
In~\cite{sun2015automatic}, unreliable concepts extracted from metadata are filtered by cross-validation average precision from a simple classifier. The remaining concepts are then clustered as concept vocabulary for downstream tasks.
\cite{berg2006animals} selects visual exemplars using a clustering method with metadata, and manually assigns concepts for cluster exemplars to guide the classifier training.
Furthermore, ConceptLearner~\cite{zhou2015conceptlearner} uses an automatic threshold method for concept allocation to clusters.
Still, based on clustering, NEIL~\cite{chen2013neil} establishes a lifelong training system that progressively identifies concepts and expands the dataset for better classifiers. 
Concept detectors and extra knowledge of concept relationships are also applied in NEIL.
Those labor-free clustering methods focus on efficiently defining reasonable concepts from web data, rather than dealing with massive semantic label noise where web labels and target concepts are mismatched. 
In this paper, we collect precise concept definitions from WordNet~\cite{miller1998wordnet}. Metadata is utilized to pinpoint reliable images with correct concepts for each category. Our work is orthogonal to concept learning. 

\subsection{Text Embedding}
Text embedding is a fundamental topic in NLP and has made considerable progress in the past decade. With the emergence of neural probability language models, word2vec becomes one of the most widely used word embedding methods. 
This method uses self-supervised representation learning, which assumes that words placed in a similar context have close meanings~\cite{mikolov2013word2vec}. 
Recent works uses powerful Transformer models~\cite{vaswani2017attention} with carefully-designed pretraining tasks and large-scale corpora, which can even outperform human performance on multiple NLP benchmarks \cite{su2019vlbert,yang2019xlnet,liu2019roberta}. In our work, we use XLNet to encode the metadata and label description into a vector format.

\subsection{Graph Neural Networks}
\label{sec:related-gnn}
Graph neural networks (GNNs) become popular in the last five years when combining deep learning and graph theory to learn from graph structured data~\cite{wu2020comprehensive,zhou2018graph}.
Representative models ChebNet~\cite{defferrard2016convolutional} and GCN~\cite{kipf2016semi} use efficient filtering approaches in graph convolution operators to improve scalability and robustness. Plentiful GNN variants emerge sequentially. The simplest GNN model, simple convolutional network (SGC)~\cite{wu2019simplifying}, claims that while eliminating unnecessary complexity and redundant computations, its simplification can still maintain accuracy.

Recent works also attempt to deal with image classification tasks on the graph settings. Some works build a knowledge graph to provide possibility of label coexistence for multi-label classification~\cite{chen2019multi}. 
Some other works do not foucs on label space, but use the neighborhood on the visual feature space to enhance the classifier~\cite{yang2019learning,iscen2019label}.
In our work, based on the observation in Figure~\ref{fig:intro_tsne} (even if semantic label noise is severe, features extracted from DNNs can still be clustered based on semantic information), we follow the visual feature graph path and train GNN for label correction.

\section{Proposed Method}
To correct the massive semantic label noise, our strategy is to use text metadata to provide correct semantic guidance for label correction, on the graph spanned in the training feature space. 
Since our label correction method is characterized by semantic guidance on the visual graph, we entitle it as a visual-semantic graph-based label corrector, referred as `VSGraph-LC'. 
The flow chart of VSGraph-LC is illustrated in Figure~\ref{fig:pipeline}. 
The procedures of VSGraph-LC are presented in an algorithm form in the appendix. 
Details of our approach is explained in the following subsections.


\begin{figure*}[t]
	\centering
	\includegraphics[width=\linewidth]{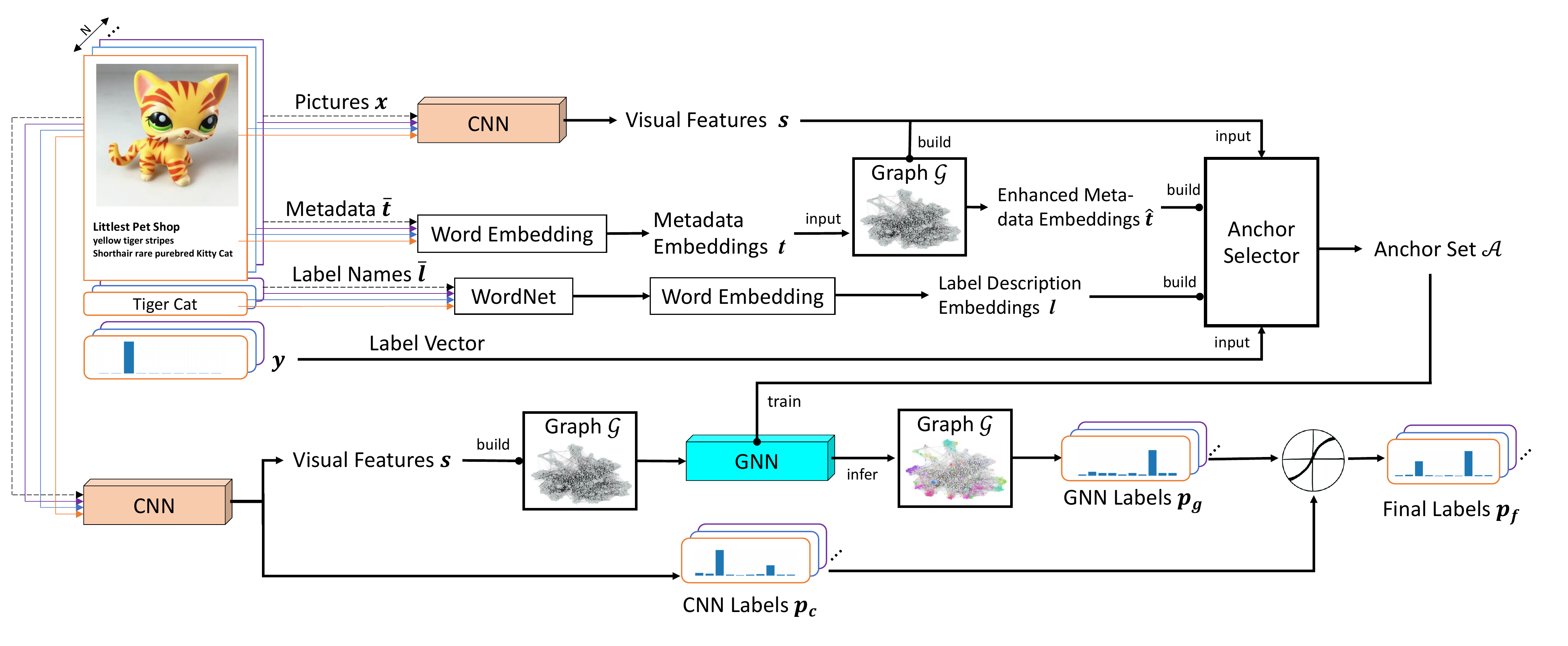}
	\caption{Pipeline of VSGraph-LC. The aim is to provide final labels that are more reliable than web labels for later finetuning. A metadata-based anchor selector is built firstly to provide guidance for GNN training. GNN and CNN labels are the predictions of GNN and the pretrained CNN model, respectively. The final labels take advantages of both CNN and GNN labels}
	\Description{The pipeline of the proposed method VSGraph-LC}
	\label{fig:pipeline}
\end{figure*}

\subsection{Problem Definition and Notations}
\label{S:problem}
Traditional webly supervised learning problems aim to train the CNN model $\mathcal{M(\theta)}$ for the optimal parameter $\hat{\theta}$ from dataset $\mathcal{D}=\{(x_1, y^*_1),\dots,(x_N, y^*_N)\}$~\cite{li2017webvision}.
Consider the massive label noise, web label $y_i^*$ might not reflect the correct category that $x_i$ belongs to~\cite{xiao2015learning}.
Thus, our task is to propose a label corrector VSGraph-LC which provides final labels $p_f$ to correct the former noisy labels $y^*$ for finetuning on the pretrained CNN model $\mathcal{M}{(\theta_0)}$, with the aid of metadata $\bar{t_i}$ crawled together with $x_i$ and $y^*_i$.
For notation, we use non-subscript style to represent the matrix that collects all vectors of the entire dataset. For example, $y^*$ represents the matrix collecting all $y^*_i$.
We also denote the matrix of label names as $\bar{l}$.

\subsection{Visual Graph Construction}
With a CNN model $\mathcal{M}{(\theta_0)}$ that pretrained on the entire webly training set $\mathcal{D}$, for every sample $x_i$, we obtain a visual feature $s_i$ that is extracted before fully-connected layer and a CNN label ${p_c}_i$ equivalent to $p(y|x_i,\theta_0)$, the prediction of $\mathcal{M}{(\theta_0)}$. With visual features $s$, we construct a k-nearest neighbor~\cite{dudani1976distance} undirected graph $\mathcal{G}=\{\mathcal{V}, \mathcal{E}\}$ where node set $\mathcal{V}$ contains every sample in the training set attached with corresponding visual and text features for later use. Information of edge set $\mathcal{E}$ is in weighted adjacency matrix $\mathbf{A}$.
\begin{equation}
\label{E:adj}
\begin{aligned}
A_{ij} =
\begin{cases}
~\cos(s_i, s_j) &, ~ \text{if}~v_i \in \mathcal{N}_k(v_j) ~\text{or}~ v_j \in \mathcal{N}_k(v_i)\\
~0 				   &, ~ \text{otherwise,}
\end{cases}
\end{aligned}
\end{equation}
where $\mathcal{N}_k(v_i)$ denotes the set of $k$ neighbors of node $v_i$ and $\cos(s_i,s_j)$ calculates the cosine similarity between two features $s_i$ and $s_j$. Till then, the visual graph is built completely.
	
\subsection{Text Embedding}
\label{S:metadata}
In this section, we prepare embeddings for unstructured text metadata and label description. After obtaining the raw metadata $\bar{t}$ from the dataset, we remove all punctuations, digits, and stop words from the raw text, followed by the tokenization, stemming, and lemmatizing. The preprocessed metadata is encoded by an off-the-shelf document embedding model. Denote the joint process of preprocessing and document embedding as function $E_{doc}$, the metadata embedding $t_i$ can be expressed as
\begin{equation}
\label{E:metadata}
t_i = E_{doc}(\bar{t_i})
\end{equation}

To encode the label concept, however, we cannot apply $E_{doc}$ directly on the label name $\bar{l_i}$ since the label name itself only contains a few words which may be non-descriptive and semantically confusing. This problem can be solved by involving a semantic knowledge base to obtain a detailed label description for a given concept. In our work, we use WordNet \cite{miller1998wordnet}, a lexical database that arranges distinct cognitive synonyms (called `synsets’) in a tree structure, to enhance the descriptive power of the label name. Similar to \cite{wei2015labeldescription}, our label description is obtained by extracting the definitions and lemma names of the original web label synset and its adjacent synsets, including hyponyms (subclass of) and member holonyms (part of). The purpose of collecting adjacent synsets is to include potentially related concepts. For instance, the label description of class `drumstick' is `drumstick: a stick used for playing a drum' for the original web synset plus `mallet, hammer: a light drumstick with a rounded head that is used to strike such percussion instruments as chimes, kettledrums, marimbas, glockenspiels, etc.' for its adjacent synsets. 

For each label name $\bar{l}_i\in \{\bar{l}_1,\dots,\bar{l}_C\}$, where $C$ is total number of categories, we locate the corresponding synset in WordNet and extract its label description, denoted as $\phi(\bar{l_i})$. Using $E_{doc}$ still, the label description embedding $l_i$ is obtained by
\begin{equation}
\label{E:label_description}
l_i = E_{doc}(\phi(\bar{l_i})).
\end{equation}

\subsection{Anchor Selection}
	\label{S:anchor}
	From t-SNE visual feature visualization~\cite{maaten2008visualizing} in Figure~\ref{fig:intro_tsne}, we observe that even under the same web label category, the CNN model is still able to cluster features according to concepts, but the discriminative function has a large bias because of semantic label noise. Thus, it is necessary to set anchors for categories to pinpoint the correct concept. We use metadata to achieve this goal.
	
	Firstly, we enhance the metadata embeddings $t$ by applying a graph smoothing function explained in~\cite{kipf2016gcn}. The enhanced metadata embeddings $\hat{t}$ are calculated as
	\begin{equation}
	\label{E:enhanced_text}
	\hat{t}=\mathbf{D}^{-\frac{1}{2}}(\mathbf{A}+w\mathbf{I})\mathbf{D}^{-\frac{1}{2}}t,
	\end{equation}
	which aggregates metadata embeddings of neighbors on the visual graph to alleviate noise from metadata. The degree matrix $\textbf{D}$ is diagonal with its element $D_{ii}=\sum_{j}A_{ij}$. Self-weight ratio $w$ is a scalar to decide the proportion of origin $t$ in the enhanced $\hat{t}$. $\textbf{I}$ is an identity matrix.
	
	For sample $i$, $l_{y^*_i}$ denotes the label description embedding of its web label name. Cosine similarity between $\hat{t_i}$ and $l_{y^*_i}$ reflects the possibility that it belongs to its web label's correct concept.
	We select $m$ anchors from each class to form anchor set $\mathcal{A}$ as Equation~\ref{E:anchor}, igniting the next GNN labeling process. $\tau_{y^*_i}^m$ equals to the $m$-th highest value in web category $y^*_i$, ensuring equal number of anchors are selected in each class.
	
	\begin{equation}
	\label{E:anchor}
	\mathcal{A}=\Big\lbrace\left(  s_i, y^*_i \right)  \Big| ~ \cos\left(\hat{t}_i, l_{y^*_i} \right) \ge \tau_{y^*_i}^m  \Big\rbrace
	\end{equation}
	
	\subsection{Graph Neural Networks Labeling}
	\label{S:gnn}
	We conduct GNN training on the graph $\mathcal{G}$. According to occam's razor~\cite{blumer1987occam}, a basic $L$-layer simple graph convolutional network (SGC)~\cite{wu2019simplifying} is implemented. For layer $i\in \{1,\dots, L\}$, the input $h^{(i-1)}$ is transformed into $h^{(i)}$ by
	\begin{equation}
	\label{E:gnn}
	h^{(i)} = \mathbf{D}^{-\frac{1}{2}}(\mathbf{A}+w\mathbf{I})\mathbf{D}^{-\frac{1}{2}}h^{(i-1)}\theta_g^{(i)},
	\end{equation}
	where $\theta_g^{(i)}$ is the corresponding trainable parameter. Hyperparameter $w$ follows Equation~\ref{E:enhanced_text}.
    Although the GNN operation is applied to the entire visual feature set, the loss is only computed on the selected anchor set $\mathcal{A}$. The first layer input $h^{(0)}$ is assigned by visual features $s$, and the final output $h^{(L)}=p(y|s,\theta_g)$, where $\theta_g$ collects all the training parameters $\theta_g^{(i)}$ across layers. The loss function is
	\begin{equation}
	\label{E:gnn_loss}
	\mathcal{L}_{g}=\sum_{(s_i,y^*_i)\in \mathcal{A}}-y^*_i\log\left(h^{(L)}_i\right).
	\end{equation}
	
	After the optimal $\hat{\theta}_g$ is obtained from training, the labeling process starts by applying the GNN inference for the entire visual feature set. The prediction $p(y|s,\hat{\theta}_g)$ is directly utilized as GNN labels $p_g$.
	
	\subsection{Correct Label Estimation}
	\label{S:final-label}
	In the experiments, we observe that GNN labels with high confidence are usually reliable samples with clean background.
	In contrast, samples with low GNN label confidence are either hard samples or open-set noise, which are difficult or impossible to be classified. To fully exploit these samples beyond the reach of GNN, CNN label $p_c$, the prediction of the pretrained CNN model, is utilized to correct low-confident GNN labels.
    
    Formally, a method for combining GNN and CNN labels is proposed as
	\begin{equation}
	\label{E:combination}
	\begin{aligned}
	{p_f}_i = 
	\begin{cases}
    {p_g}_i & ,~ \max({p_g}_i)\ge\tau_f\\
	\lambda {p_g}_i + (1-\lambda){p_c}_i & ,~ \text{otherwise,}
	\end{cases}
	\end{aligned}
	\end{equation}
	
	where $\lambda$ controls the contribution of CNN labels when GNN labels have lower confidence scores than threshold $\tau_f$.
    
    \subsection{Summary}
    In summary, our VSGraph-LC firstly selects anchors based on metadata and label description. Anchors propagate their correct semantic concepts across the visual feature graph. 
    The process of VSGraph-LC finishes when the final label $p_f$ is generated. With the corrected label, we finetune the pretrained CNN model with loss function
    
	\begin{equation}
	\label{E:cnn_loss}
	\mathcal{L}_c=\sum_{(x_i,{p_f}_i)\in \mathcal{D}}-{p_f}_i \log\left(  p(y|x_i,\theta)  \right).
	\end{equation}

\section{Experiments}
In this section, we evaluate our method VSGraph-LC on WebVision-1000~\cite{li2017webvision} and NUS-WIDE~\cite{nus-wide-civr09}. To accelerate the experiments, we generate a compact Google-500 dataset from massive WebVision-1000 for ablation study and discussion. Our method reaches the state-of-the-art result on WebVision-1000 and proves its robustness and generalization on noisy multi-label dataset NUS-WIDE. Besides, we investigate the progressive training strategy for VSGraph-LC in Section \ref{S:iterative}. We also find our method more powerful on open-set validation set in Section ~\ref{S:openset}.

\subsection{Datasets, Metadata and Configurations}
\label{S:Dataset}
\paragraph{WebVision-1000} WebVision-1000~\cite{li2017webvision} contains $2.4M$ web images crawled from Flickr and Google, with keywords from $1000$ class-labels in ILSVRC-2012~\cite{deng2009imagenet} (known as ImageNet).
The estimated top-1 label accuracy is $48\%$~\cite{guo2018curriculumnet}. 
As WebVision shares the same 1000 classes with ImageNet, we also use ImageNet validation set along with WebVision-1000's own validation set for evaluation.

For the WebVision dataset, every training sample contains extra attributes crawled along with the image, including the rank in searching result, title, description, tags, source website, etc. Considering both cleanness and descriptiveness, we choose `title' + `description' attributes for Google images and `title' + `tags' attributes for Flickr images as the metadata.

\paragraph{Google-500} Google-500 only keeps images crawled from google websites in WebVision-1000 for their more completed and cleaner metadata than Flickr. Also, we randomly sample one-half categories to alleviate the large consumption of time and GPU resources without losing generalization with a total of 489755 samples. Validation sets of selected categories remain. We mainly use Google-500 for ablation studies.
The metadata details refer to WebVision-1000. 

\paragraph{NUS-WIDE} NUS-WIDE~\cite{nus-wide-civr09} is a real-world web image dataset that contains 269,648 images with the total number of 5018 associated tags crawled from Flickr. Each image contains web tags and human-annotated ground-truth labels for the 81 concepts. Since the dataset does not contain the original web queries, we obtain web labels by extracting labels from images’ associated tags among these 81 labels, i.e. we check whether each of the 81 labels appears in its web tags. It is reported in \cite{nus-wide-civr09} that on average 50\% of the web labels are incorrect and 50\% of the ground-truth labels are missing in web labels. All web tags for an image are used as its metadata. 

\begin{figure*}[t]
	\centering
	\begin{subfigure}{\textwidth}
		\centering
		\includegraphics[width=\linewidth]{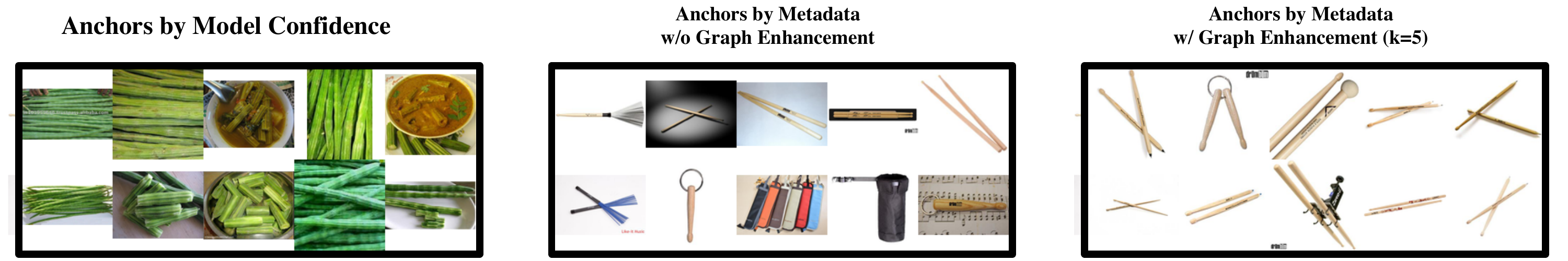}  
		\caption{Selected Anchors for Class `Drumstick'}
		\Description{Selected Anchors for Class `Drumstick'}
		\label{fig:anchor_drumstick}
	\end{subfigure}
	\begin{subfigure}{\textwidth}
		\centering
		\includegraphics[width=\linewidth]{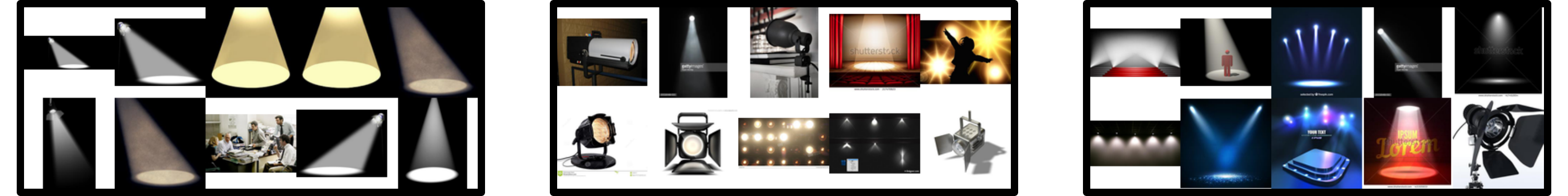}  
		\caption{Selected Anchors for Class `Spotlight'}
		\Description{Selected Anchors for Class `Spotlight'}
		\label{fig:anchor_spotlight}
	\end{subfigure}
	
	\begin{subfigure}{\textwidth}
		\centering
		\includegraphics[width=\linewidth]{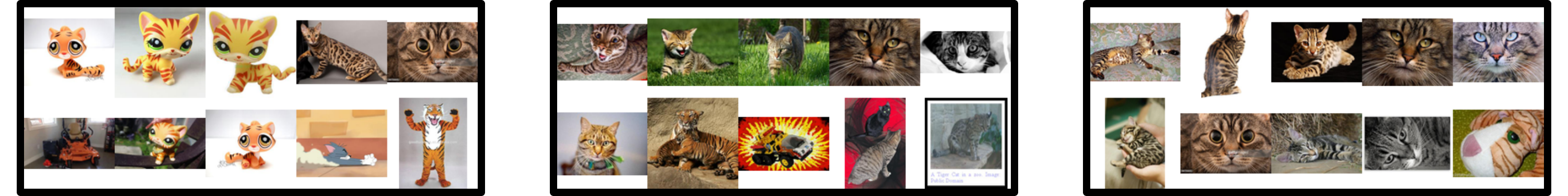}  
		\caption{Selected Anchors for Class `Tiger Cat'}
		\Description{Selected Anchors for Class `Tiger Cat'}
		\label{fig:anchor_tigercat}
	\end{subfigure}
	
	
	
	\caption{Exemplar anchors for three most noisy classes in Google-500. Different columns indicate anchors selected by different methods. Using metadata with graph enhancement has perceptible advantages compared to other methods}
	\Description{Exemplar anchors for five most noisy classes in Google-500}
	\label{fig:seed}
\end{figure*}

\paragraph{Configuration details.}
ResNet50~\cite{he2016deep} is selected as our CNN model in all experiments.
For all experiments, we set batch size as 256 and mini-batch size as 32 trained on 8 GPUs, except in WebVision-1000 we set batch size as 1024 on 32 GPUs. We use the standard SGD with the momentum of $0.9$ and weight decay of $10^{-4}$. A warm-start linearly reaches the initial learning rate in the first $10$ epochs. The remained epochs are ruled by a cosine learning rate scheduler. A simple class reweighting is performed to deal with class imbalance. Google-500/WebVision-1000 requires training from scratch with $120$/$150$ epochs and an initial learning rate of $0.1$/$0.4$. 
In the finetuning stage, initial learning rate is cut half from origin one without warm-start.
Training on NUS-WIDE requires an ImageNet pretrained model with $100$ epochs and learning rate $0.002$.
For the GNN model, we take the 1-layer SGC model with learning rate of $0.1$, $5000$ epochs and Adam optimizer with weight decay of $10^{-6}$.

\subsection{Google-500}
\label{sec:google-500}
In this section, we experiment VSGraph-LC on the Google-500 dataset. We first evaluate the anchor selector and scrutinize the quality of selected anchors in Section~\ref{S:exp-anchor}. We then confirm that the final label calculation policy in Section~\ref{S:final-label} can achieve a better result than only using any one of the component labels. Finally, we evaluate the performance of VSGraph-LC under various hyperparameters, showing its robustness.

\subsubsection{Anchor Selector}
\label{S:exp-anchor}
With metadata embeddings $t$ and label description embeddings $l$ well prepared, anchor selector will be built according to Section~\ref{S:anchor}. Firstly we choose $k=5$ for $k$-NN graph $\mathcal{G}$ and self weight $w=0$ for text feature aggregation, which means the enhanced text feature only depends on neighbours' embeddings. The selection of hyperparameter will be explained in Section~\ref{S:exp-hyper}.
Figure~$\ref{fig:seed}$ shows anchors from 3 classes with lowest classification accuracy according to pretrained model $\mathcal{M}(\theta_0)$. Top 10 anchors selected by different methods are shown. The visualization shows that $\mathcal{M}(\theta_0)$ predicts high confidence on samples with an incorrect semantic concept regarding class `drumstick' and `tiger cat', which reflects that $\mathcal{M}(\theta_0)$ is misled by massive semantic label noise from these classes. Anchors selected by metadata without graph enhancement can also make mistakes. Fortunately, when the graph-enhanced text features are introduced, those mistakes made by isolated metadata are largely mitigated by insurance from samples' neighborhood. 
In the following experiments, we keep $m=10$ for the Google-500 anchor selector and expect the selected anchors to provide reliable guidance for the graph neural network.

\subsubsection{GNN Training, Final Labeling and Results.}
With graph $\mathcal{G}$ constructed, simple graph convolutional network (SGC)~\cite{wu2019simplifying} is utilized for training followed by Section~\ref{S:gnn}. Table~\ref{tab:google500} shows the deficient performance if we only use the GNN label as the final label for finetune. As explained in Section~\ref{S:final-label}, with GNN, anchors will only propagate their labels to nearby clean and easy samples on the graph, whereas for those hard samples or open-set noise, GNN prefers to give them very low prediction scores, acting like dropping numerous data that harm the data-driven DNNs. Thus, we set $\lambda=0.5$ for Equation~$\ref{E:combination}$, therefore the model finetuned by the final label will have a large improvement than the pretrained model $\mathcal{M}{(\theta_0)}$ and the model finetuned by only CNN or GNN labels. The results also have advantages over Co-teaching.

\setlength{\tabcolsep}{4pt}
\begin{table}
	\centering
	\caption{Ablation study on Google-500 dataset}
	\label{tab:google500}
	\begin{tabular}{lcccc}
		\toprule
		\multirow{2}{*}{~ Method}  & \multicolumn{2}{c}{ ~ WebVision ~ } & \multicolumn{2}{c}{ ~ ImageNet ~ } \\
		\multicolumn{1}{c}{} & \multicolumn{1}{c}{ ~ Top-1 ~ } & \multicolumn{1}{c}{ ~ Top-5 ~ } & \multicolumn{1}{c}{ ~ Top-1 ~ } & \multicolumn{1}{c}{ ~ Top-5 ~ } \\
		\midrule
		Pretrained model &66.96 &82.68&61.54 &78.89\\
		\midrule
		Co-teaching  & 67.61 & 84.04 & 62.18 & 80.98 \\
		\midrule
		Finetune by $p_g$ only &64.79&81.22	&60.39&78.76\\
		Finetune by $p_c$ only &67.83&83.93	&62.68&80.68\\
		Finetune by $p_f$      &\textbf{68.14}&\textbf{84.46}&\textbf{63.16}&\textbf{81.45}\\
		\bottomrule
	\end{tabular}
\end{table}

\subsubsection{Hyperparameters}
\label{S:exp-hyper}
\begin{figure}[t]
	\centering
	\begin{subfigure}{.23\textwidth}
		\centering
		\includegraphics[width=\linewidth]{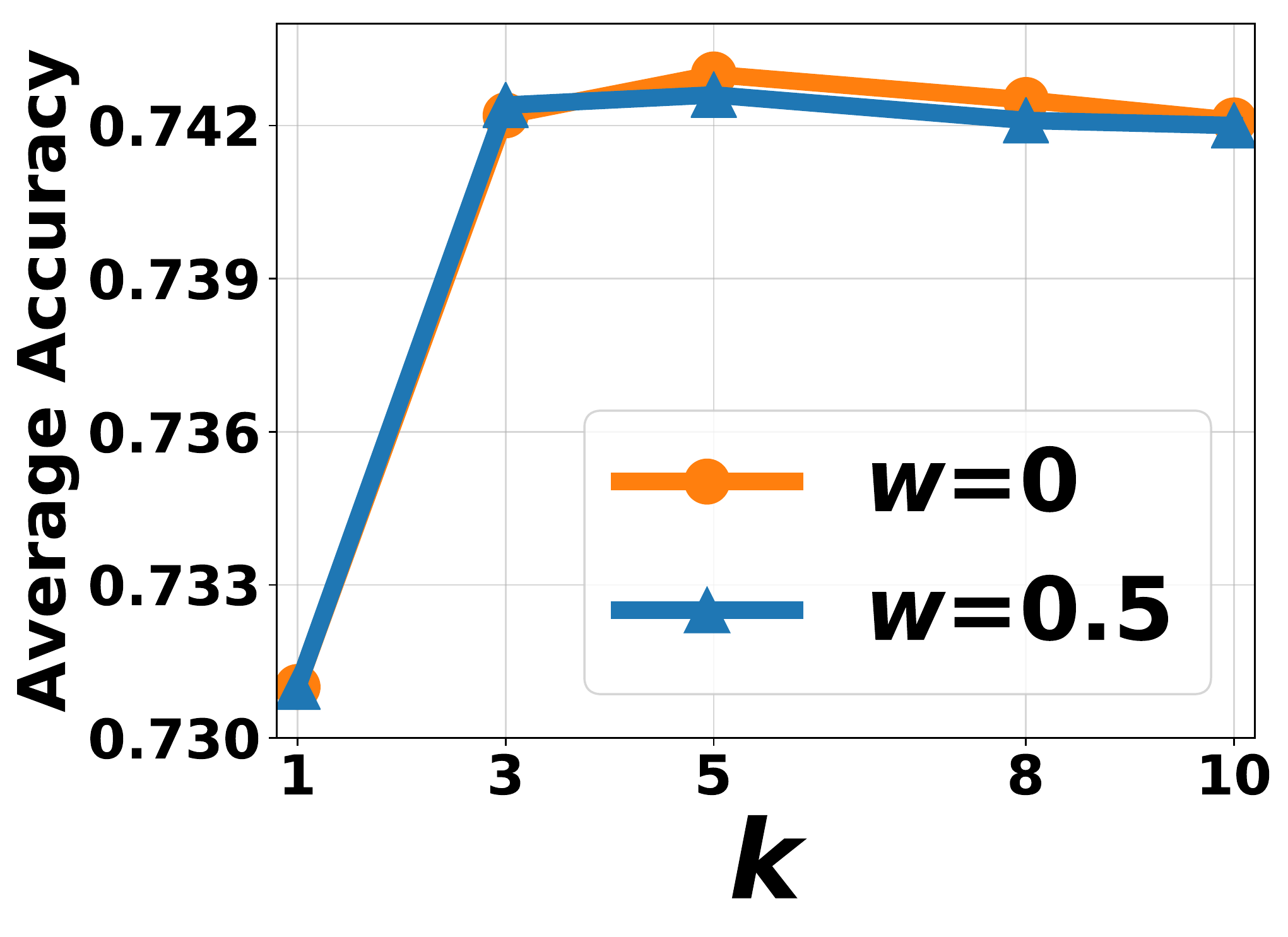}  
		\caption{Avg. acc. with $k$ and $w$}
		\Description{Average accuracy with $k$ and $w$}
		\label{fig:hyper1}
	\end{subfigure}
	\begin{subfigure}{.23\textwidth}
		\centering
		\includegraphics[width=\linewidth]{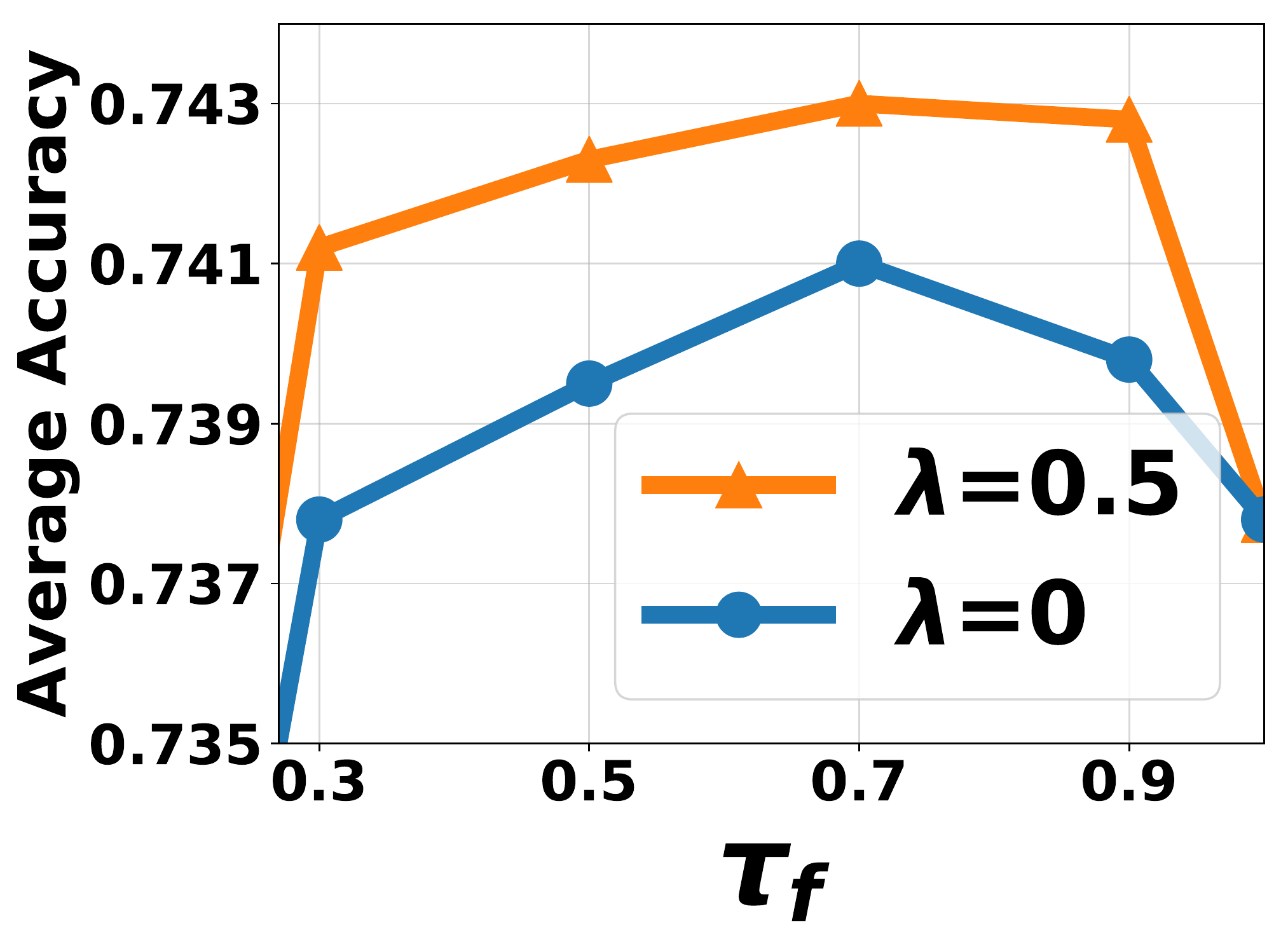}  
		\caption{Avg. acc. with $\lambda$ and $\tau_f$}
		\Description{Average accuracy with $\lambda$ and $\tau_f$}
		\label{fig:hyper2}
	\end{subfigure}
	\caption{Comparison between different hyperparameters}
	\Description{Comparison between different hyperparameters}
	\label{fig:hyper}
\end{figure}

In this section, we try several values for critical hyperparameters of $k$, $w$ for k-NN graph, and $\lambda$, $\tau_f$ for final label calculation in Equation~\ref{E:combination}. 
Mean values of WebVision/ImageNet top-1/top-5 are used as average accuracy to indicate the overall performance. Figure~\ref{fig:hyper1} shows the effectiveness of using k-NN, where $k=5$ ensures an optimal result. Figure~\ref{fig:hyper2} shows $\lambda=0.5$ has constant merit comparing to $\lambda=0$. $\tau_f=0.7$ can assist to achieve optimal average accuracy. 
However, although the advantages of selected hyperparameters, the accuracy difference does not vary much for $\tau_f\in[0.5,0.9]$, $k\in[3,8]$ and any $w$, showing the robustness of proposed VSGraph-LC to hyperparameters.

\subsection{WebVision-1000}

Table~\ref{tab:web1000} reports experimental results on WebVision-1000. VSGraph-LC gains a large improvement on WebVision top-1 by more than $1.2\%$. The advantage is even larger on ImageNet, especially on ImageNet top-5, showing a good generalization ability. In comparison, finetuning by CNN labels only obtains weaker improvements.

Notice that our method also outperforms the state-of-the-art methods on WebVision and ImageNet validation sets. Except CurriculumNet, MentorNet and CleanNet both adopt extra human-verified datasets to train a guidance network first, while MentorNet chooses a backbone of InceptionResNetV2 which is stronger than our ResNet50. With these disadvantages, however, our method can still obtain a better performance compared to the above methods. In addition, Multimodal uses ImageNet data for training visual embedding and a query-image pairs dataset for training phrased generation, with stronger InceptionV3 being the backbone. Our VSGraph-LC can still exceed Multimodel in WebVision top-1 accuracy. 

\setlength{\tabcolsep}{1pt}
\begin{table}
	\centering
	\caption{The state-of-the-art results on WebVision-1000}
	\label{tab:web1000}
	\begin{tabular}{llcccc}
		\toprule
		\multicolumn{1}{l}{\multirow{2}{*}{~ Method}}  & \multicolumn{1}{l}{\multirow{2}{*}{Backbone}} & \multicolumn{2}{c}{WebVision} & \multicolumn{2}{c}{ImageNet} \\
		& \multicolumn{1}{c}{} & \multicolumn{1}{c}{Top-1} & \multicolumn{1}{c}{Top-5} & \multicolumn{1}{c}{Top-1} & \multicolumn{1}{c}{Top-5} \\
		\midrule
		~ MentorNet \cite{jiang2017mentornet} &InceptionResNetV2 ~ &72.60&88.90&64.20&84.80\\
		~ CleanNet \cite{lee2018cleannet} &ResNet50&70.31&87.77&63.42&84.59\\
		~ CurriculumNet \cite{guo2018curriculumnet} ~ &InceptionV2&72.10&89.20&64.80&84.90\\
		~ Multimodal \cite{shah2019inferring} & InceptionV3&73.15&89.73&-&-\\
		\midrule
		~ Pretrained model      ~ & ResNet50 &74.25&89.84&68.28&86.23\\
		~ Finetune by $p_c$ only~ & ResNet50 &75.15&89.93&69.07&86.76\\
		~ Finetune by $p_f$     ~ & ResNet50 &\textbf{75.48}&\textbf{90.15}&\textbf{69.42}&\textbf{87.29}\\
		\bottomrule
	\end{tabular}
\end{table}

\subsection{NUS-WIDE}
NUS-WIDE~\cite{nus-wide-civr09} is another real-world web image dataset. Different from WebVision, NUS-WIDE is designed for multi-label image classification.
Previous works~\cite{Wang2016cnnrnn,Zhu_2017_CVPR} only use the dataset with ground-truth labels for standard multi-label learning, while we are interested in the real-world label noise setting in multi-label learning. In this case, we train our model with web labels and evaluate it on ground-truth labels.
To distinguish from previous multi-label classifier learned without noise, we denote their and our setting as NUS-81 and NUS-81-Web, respectively. The only difference between NUS-81 and NUS-81-Web is that the NUS-81-Web training set uses web labels rather than annotated ground-truth labels. The train-test split policy adopts the official one.
For experiments, we remove samples without any label within the 81 label set. Moreover, the label descriptions are obtained by identifying the most relevant synset on WordNet for each label.


\setlength{\tabcolsep}{4pt}
\begin{table}
	\centering
	\caption{Results on NUS-81-Web with noisy web labels for training. $K=3$ is used for calculating C-F1 and O-F1}
	\label{tab:nuswide}
	\begin{tabular}{lcccc}
		\toprule 
		Method & C-F1 & O-F1 & mAP \\
		\midrule
		~ Pretrained model ~  &37.51 &39.59 &43.94\\
		~ Finetune by $p_c$ only~ &37.62 &39.15 &43.99 \\
		~ Finetune by $p_f$~ & \textbf{38.58} & \textbf{40.16} &\textbf{44.83}\\
		\bottomrule
	\end{tabular}
\end{table}

\subsubsection{Evaluation Metrics}
For multi-label classification, we compare the overall F1-measure (O-F1), per-class (also known as macro-averaged) F1-measure (C-F1), and mean average precision (mAP) to evaluate the performance.
Following~\cite{Wang2016cnnrnn}, we use top $K=3$ highest confidence labels for each image as the prediction and compare with the ground-truth labels. 

\subsubsection{Anchor Selector and GNN Training}
With the high noisy ratio and small dataset size for NUS-81-Web, the model trained directly from scratch using web labels cannot provide strong visual features for building $k$-NN graph. 
Therefore, we finetune the ImageNet-pretrained model for our pretrained model $\mathcal{M}(\theta_0)$ instead. Regarding hyperparameters, we set $k=10$ for $k$-NN graph, self-weight $w=0$ for text enhancement and $m=50$ for anchor selector.  

\subsubsection{Results}
Experimental results on NUS-81-Web are shown in Table~\ref{tab:nuswide}. Our proposed method outperforms baseline model and the model finetuned by only CNN labels for all three metrics.
For the mAP score, we achieve $0.9\%$ improvement using VSGraph-LC compared to the pretrained model. Our method can also increase both C-F1 and O-F1 by $1.1\%$ and $0.6\%$, respectively.
Unlike the performance on WebVision, finetuning with CNN labels brings no significant improvement on C-F1 and mAP, and even a $0.4\%$ drop on O-F1. 
This demonstrates that under a high noisy-level setting like NUS-81-Web, labels corrected by our VSGraph-LC method are more reliable than the CNN labels.

\setlength{\tabcolsep}{4pt}
\begin{table}
	\centering
	\caption{Results on Google-500 open-set problem}
	\label{tab:openset}
	\begin{tabular}{lcccc}
		\toprule
		\multirow{2}{*}{~ Method}  & \multicolumn{2}{c}{WebVision} & \multicolumn{2}{c}{ImageNet} \\
		\multicolumn{1}{c}{} & \multicolumn{1}{c}{C-P/C-R} & \multicolumn{1}{c}{C-F1}  & \multicolumn{1}{c}{C-P/C-R} & \multicolumn{1}{c}{C-F1}  \\
		\midrule
		Pretrained model    &40.44/67.23	& 50.50 & 37.75/60.36 & 46.45\\
		Co-teaching         &40.38/69.21    & 51.00 & 38.27/63.58 & 47.78\\
		Finetune by $p_c$   &40.34/68.21    & 50.70 & 38.54/62.10 & 47.56\\
		Finetune by $p_g$   &59.64/55.63	& \textbf{57.56} & 56.74/48.89  &52.52\\
		Finetune by $p_f$   &52.47/62.95	&57.24 &49.66/56.36	& \textbf{52.80}\\
		\bottomrule
	\end{tabular}
\end{table}
\setlength{\tabcolsep}{1.4pt}

\subsection{Discussion on Progressive Training}
\label{S:iterative}
\begin{figure}[t]
	\centering
	\begin{subfigure}{.22\textwidth}
		\centering
		\includegraphics[width=\linewidth]{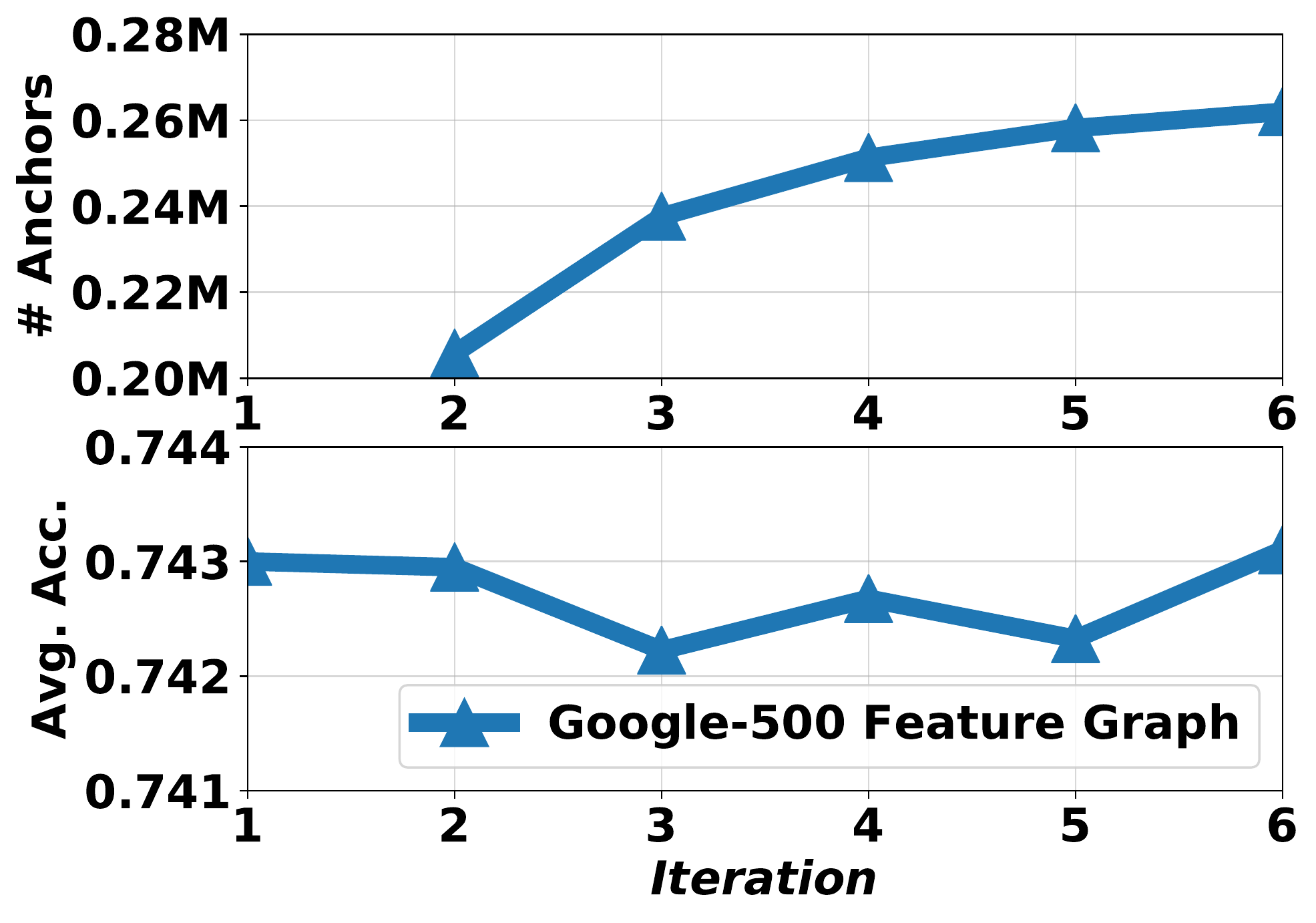}  
		\caption{Google-500 Feature Graph}
		\Description{Google-500 Feature Graph}
		\label{fig:iter1}
	\end{subfigure}
	\begin{subfigure}{.22\textwidth}
		\centering
		\includegraphics[width=\linewidth]{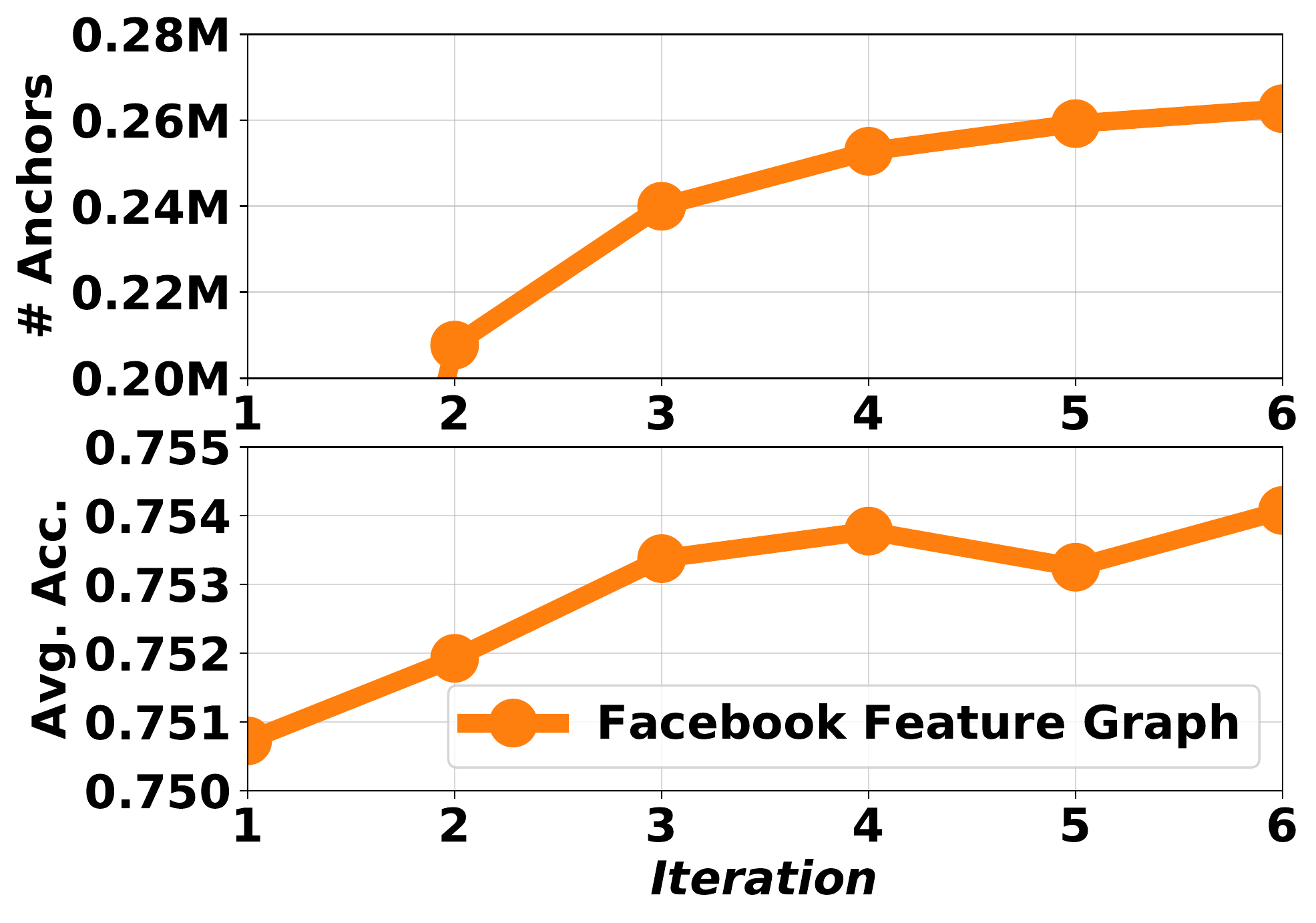}  
		\caption{Off-the-Shelf Feature Graph}
		\Description{Off-the-Shelf Feature Graph}
		\label{fig:iter2}
	\end{subfigure}
	\caption{Exploration on progressive training, which only works with graph built by high-quality visual features}
	\Description{Exploration on progressive training, which only works with graph built by high-quality visual features}
	\label{fig:iter}
\end{figure}

In this section, we discuss the performance of progressive GNN training. The procedure is as follows:
After the standard GNN training is completed as Section~\ref{S:gnn}, samples with confidence over threshold $\tau_f$ will be selected as anchors, labeled by GNN labels, for GNN training in the next round. 
Progressive GNN training utilizes the static graph that built by the pretrained CNN features. We suppose that with increasing iterations, the performance of VSGraph-LC will keep growing, boosted by more anchors.
However, Figure~\ref{fig:iter1} shows that more iterations cannot guarantee a better performance under Google-500 settings in Section~\ref{sec:google-500}, even though the number of anchors increases according to top Figure~$\ref{fig:iter1}$.
We assume that the failure attributes to the quality of graph structure $\mathcal{G}$ built by purely Google-500 base model features. Unable to build the connection between easy and hard samples of the identical category, the existing graph $\mathcal{G}$ only constrains GNN labels within easy samples, thus disables progressive training.
To prove this, we build the visual graph using strong features extracted by an off-the-shelf ResNeXt-101 model provided by~\cite{yalniz2019billion}, which is trained on more than 940 million images in semi-weakly supervised learning fashion.
We evaluate progressive training again by ONLY changing the graph structure, keeping the original 5000 anchors and all node features and scores the same. 
In this case, initial anchors can reach hard and more informative samples, accumulated as new anchors for progressive training. During the iteration, $\mathcal{G}$ remains unchanged as before.
The results show a stable improvement iteration by iteration, indicating the importance of visual graph structure for progressive training of VSGraph-LC.

\subsection{Discussion on Open-Set Recognition (OSR)}
\label{S:openset}
Due to the complexity of real-world scenarios, DNN is ideally resistant to open-set images, \textit{i.e.}, for test images which do not belong to any category in the validation set, CNN models should produce low confidence for them.
To demonstrate our model's superiority on open-set recognition task, we use the final model trained from the Google-500 training set and evaluate it on the entire WebVision/ImageNet validation dataset. 
To be specific, since the WebVision/ImageNet validation sets share the same 1000 classes, we set these 1000 classes except Google-500's classes as open-set classes (i.e. 500 for classification and other 500 as open set). 
OSR expects unconfident predictions for open-set images. 
If CNN models predict an image with confidence beyond a threshold ($0.2$ in our paper), we consider it is classified into the predicted category, followed by~\cite{perera2019ocgan,yoshihashi2019classification}. 
Thus, those open-set images might be falsely classified into Google-500's classes. 
We take per-class precision (C-R), recall (C-R), and F1-measure (C-F1) as metrics in Table~\ref{tab:openset}, where VSGraph-LC has tremendous improvements than the pretrained model and Co-teaching, showing our model fits the real-world open-set tasks well.

We also find that finetuning by $p_c$ does not work in OSR as it tends to give every sample high confidence even for open-set images.
However, $p_g$ from GNN only gives high confidence to samples near anchors, thus can reject open-set samples and outperforms $p_c$ on OSR. But GNN might miss some hard positive samples, harming performance in the close-set setting. We therefore introduce $p_f$ to combine $p_c$ and $p_g$, whose OSR ability, in some cases, might be weaker than only using $p_g$ alone because of $p_c$.
    
\section{Conclusions}
In this paper, we focus on webly supervised learning task and highlight two understudied but critical factors: semantic label noise and text metadata.
Based on our extensive exploration on them, we gain insights that CNN model that pretrained from entire webly dataset is able to provide a visual feature space where similar semantic images cluster themselves. With efficient usage of metadata, an effective and automatic label corrector VSGraph-LC is proposed.

\begin{acks}
The work described in this paper was partially supported by Innovation and Technology Commission of the Hong Kong Special Administrative Region, China (Enterprise Support Scheme under the Innovation and Technology Fund B/E030/18).
\end{acks}
\newpage

\bibliographystyle{ACM-Reference-Format}
\bibliography{sample-base}

\newpage
\appendix

\section{Algorithm}
\begin{algorithm}[]
	\caption{VSGraph-LC (depicted as Figure~3)}
	\label{alg:algorithm}
	\begin{flushleft}
		\textbf{Input}: Dataset $\mathcal{D}=\{x,y^*\}$, raw metadata $\bar{t}$ for every sample, raw label names $\bar{l}$, number of anchors per class $m$.\\
		\textbf{Output}: Final corrected labels $p_f$ for finetuning $\mathcal{M}(\theta_0)$.
	\end{flushleft}
	\begin{algorithmic}[1] 
		\STATEx \hspace{-0.6cm} \textit{$\triangleright$ Prepare Word Embeddings}
		\STATE Obtain word embeddings $t$ from raw metadata $\bar{t}$ by Eq.2.
		\STATE Obtain word embeddings $l$ from label names $\bar{l}$ by Eq.3.
		
		\STATEx \hspace{-0.6cm} \textit{$\triangleright$ Obtain CNN Labels}
		\STATE Obtain pretrained CNN model $\mathcal{M}(\theta_0)$ from the entire $\{x,y^*\}$.
		\STATE Obtain visual features $s$ and CNN labels $p_c$ from $\mathcal{M}(\theta_0)$.
		
		\STATEx \hspace{-0.6cm} \textit{$\triangleright$ Obtain GNN Labels}
		\STATE Build graph $\mathcal{G}$ using $s$, with adjacency matrix $\mathbf{A}$ by Eq.1.
		\STATE Obtain graph-enhanced text features $\hat{t}$ by Eq.4.
		\STATE Generate anchor selector by Eq.5 with $m$, obtain anchor set $\mathcal{A}$.
		\STATE Train GNN on $\mathcal{G}$ supervised by $\mathcal{A}$ for optimal $\hat{\theta}_g$.
		\STATE Serve GNN prediction $p(y|s,\hat{\theta}_g)$ as GNN labels $p_g$.
		\STATEx \hspace{-0.6cm} \textit{$\triangleright$ Obtain Final Labels for Finetuning}
		\STATE Obtain final pseudo labels $p_f$ by Eq.8.
	\end{algorithmic}
\end{algorithm}

\end{document}